# A Human Action Descriptor Based on Motion Coordination

Pietro Falco, Matteo Saveriano, Eka Gibran Hasany, Nicholas H. Kirk, and Dongheui Lee

*Abstract*—In this paper, we present a descriptor for human whole-body actions based on motion coordination. We exploit the principle, well known in neuromechanics, that humans move their joints in a coordinated fashion. Our coordination-based descriptor (CODE) is computed by two main steps. The first step is to identify the most informative joints which characterize the motion. The second step enriches the descriptor considering minimum and maximum joint velocities and the correlations between the most informative joints. In order to compute the distances between action descriptors, we propose a novel correlation-based similarity measure. The performance of CODE is tested on two public datasets, namely HDM05 and Berkeley MHAD, and compared with state-of-the-art approaches, showing recognition results.

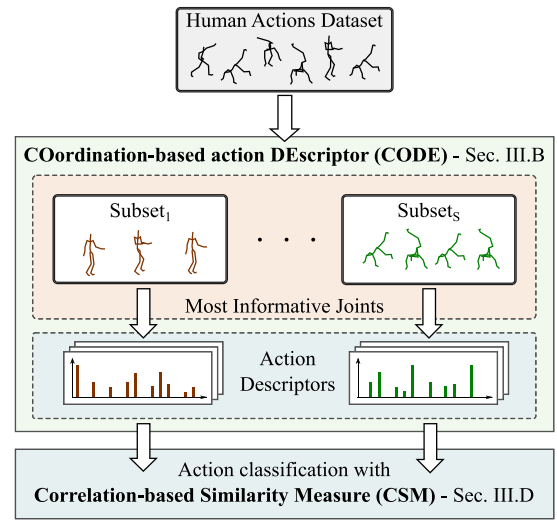

Fig. 1. Overview of the proposed approach for action representation and recognition. Intuitively, we can say that selecting the most informative joints splits the dataset into several action subsets. The actions within each subset have similar most informative joints. Neuromechanically-sound features are then added to make action descriptors more distinctive. Finally, action classification is performed using the proposed CSM metric.

## I. INTRODUCTION

IN THE last two decades, encoding and classifying human actions has been a key topic in computer vision and human movement science. Recently, motion interpretation has become a topic of great interest also within the robotic community. One of the challenges in modern robotics is to bring robots out of the structured industrial environments and let them work in close cooperation with humans. Robots will execute tasks in environments dwelled by humans and in direct contact with them. In order for robots to successfully interact with human beings, a necessary step is representing and classifying actions performed by humans.

In robotic applications, motion descriptors need to fulfill specific requirements of computational complexity and scalability in addition to accuracy. Modern autonomous robots have complex software architectures and very demanding planning and control algorithms. In order to make these systems usable in real world scenarios, it is essential to keep as low as possible the computational complexity, both for sake of time and energy consumption. Scalability is also an important issue, since in robotic applications the total number of actions and the duration of each action cannot be accurately predicted.

In order to take a step in matching these requirements, we propose a COordination-based action DEscriptor (CODE). CODE is characterized by a low time and space complexity, and achieves good scalability and classification accuracy. The concept of the proposed approach is shown in Fig. 1. CODE leverages the property of human motion, well-known in neuromechanics, that humans move their joints in a coordinated fashion [1]–[4] and that the various degrees of freedom present couplings and dependencies [2], [3]. One of the main contributions of this work is to exploit such correlation among joints to increase the performance of action recognition in terms of accuracy and scalability.

In CODE, therefore, information about correlation is a key tool to characterize motion. In order to reduce the computational complexity, CODE analyzes the correlation properties of a subset of joints, called most informative joints [5]. Roughly speaking, the most informative joints are the joints which mostly contribute to the execution of a certain action. CODE selects the most informative joints on the basis of the signal variances.

Manuscript received September 10, 2016; accepted January 4, 2017. Date of publication January 16, 2017; date of current version February 2, 2017. This paper was recommended for publication by Associate Editor K. Harada and Editor T. Asfour upon evaluation of the reviewers' comments. This work was supported in part by the Marie Skłodowska-Curie Indivisual Fellowship LEA-CON, EU project 659265, and in part by the Technical University of Munich, International Graduate School of Science and Engineering.

P. Falco, M. Saveriano, N. H. Kirk, and D. Lee are with the Chair of Automatic Control Engineering, Technical University of Munich, Munich 80333, Germany (e-mail: pietro.falco@tum.de; matteo.saveriano@tum.de; nicholas.kirk@tum.de; dhlee@tum.de).

E. G. Hasany is with the Department of Informatics, Technical University of Munich, Munich 80333, Germany (e-mail: eka.hasany@tum.de).

Color versions of one or more of the figures in this letter are available online at http://ieeexplore.ieee.org.

Digital Object Identifier 10.1109/LRA.2017.2652494

In the literature concerning motion analysis, this assumption has been proven to be valid [5] for classification applications. To increase the discriminativeness of each action, we enrich the descriptor with information about motion coordination (correlation between joint pairs), and information about velocities to discriminate the directionality of motion. Moreover, we propose a novel similarity measure, called Correlation-based Similarity Measure (CSM), which performs better than the classical Euclidean and Manhattan distances with a reduced number of informative joints.

The rest of the paper is organized as follows. Section II presents the related work. Section III describes the proposed action descriptor and similarity measure. Experiments on two human action datasets, namely Hochschule Der Medien 2005 (HDM05) [6] and Berkeley Multimodal Human Action Database (MHAD) [7], and a comparison with state-of-the-art approaches are shown in Section IV. Section V states conclusions and proposes future extensions.

## II. RELATED WORK

In the literature, there are diverse works on motion recognition, which are based on different types of input data. Two common representations of human motion are based on normalized joint positions and on joint angles. In Cartesian-based representations, motion is described with the positions of the joints in the 3D space expressed in a reference frame fixed to the human torso. As a consequence, a precise skeletal model is required for this representation. Representations based on joint angles, instead, are natively independent from the used reference frame [5]. Joint angles can be computed by inverse kinematics of a skeleton model or measured by wearable sensors such as inertial measurement units [8], [9]. This representation is potentially more interesting in robotics, since it does not implicitly assume the knowledge of the skeletal model and it does not require a normalization step. CODE is designed for angle-based representations, since the neuromechanical properties of human motion coordination have been proven for joint angles [1], [2].

*Methods Based on Joint Cartesian Positions:* Cartesian trajectories are strongly affected by the choice of the reference frame and the link lengths, which reduces the discriminative power of Cartesian descriptors [10]–[12]. To alleviate this problem, a normalization procedure is performed [10], which expresses the joint positions in a frame fixed to the torso and normalizes the length of the bones. The method is defined skeleton-based (or model-based) because it requires the knowledge of the skeletal model of the performer to obtain a user-independent normalized representation. Using this skeleton-based representation, in [10] a deep neural network is proposed to classify motion capture sequences. In [13], a hierarchical recurrent neural network is proposed for action classification. A template-based approach to recognize actions [14] uses a small set of a-priori known actions called templates. To align observed actions with the templates, the dynamic time warping [15] is adopted. In [16], a local skeleton descriptor is proposed that encodes the relative position of joint quadruples. The input data are joint Cartesian coordinates. The approach in [17] exploits learned models to represent each action and to capture the intra-class variance. The method shows promising results in dealing with data from depth cameras. The work in [18] describes a representation based on pairwise joint-to-joint distances in the skeletal model and principal component analysis is used to reduce the dimensionality.

*Methods Based on Joint Angles:* In [19], an online segmentation and recognition of manipulation task, based on singular value decomposition, is proposed. An unsupervised approach that exploits hidden Markov models to segment and recognize actions is presented in [20]. The work presented in [21] leverages the properties of human motion in frequency domain to derive a compact action descriptor. Linear Dynamical Systems (LDS) are used in [22] to recognize human gaits, and the methodology can be applied also to recognition of whole-body actions. In [5], the authors propose three descriptors ranking the most informative joints involved in an action, i.e. the joints which have highest variance during the motion. The descriptors are called Sequence of the Most Informative Joints (SMIJ), Histograms of Most Informative Joints (HMIJ) and Histogram of Motion Words (HMW), respectively. This approach is particularly significant for our work, since it proposes descriptors effective in discriminate actions but also simple and computationally efficient. This philosophy is also used in CODE as well as the concept of choosing the most informative joints based on the variance. There are two main differences between SMIJ [5] and CODE. First, CODE computes the variance of the overall motion trajectory (global descriptor) and has a constant size, while SMIJ requires to split each action into several segments (local descriptor). Second, we explicitly take into account motion coordination and propose a novel Correlation-based Similarity Measure (CSM) to compute the similarity between action descriptors. Recognition performance of LDS, HMW, SMIJ, HMIJ and CODE are compared in Section IV-D.

Aforementioned angle-based representations present two important open points. First, they are tested only on a limited set of classes (10–15 classes) and, therefore the scalability is not investigated. Second, the complexity analysis is usually neglected, even though it is an important theoretical foundation for real applicability. CODE, on the other hand, offers a good balance between accuracy, scalability, and computational complexity. CODE performs well not only on a typical datasets of 10–15 classes, but also on the whole HDM05 dataset, constituted by 80 classes and 2337 actions.

## III. PROPOSED APPROACH

This section discusses three problems related to action classification: *i)* which raw data from tracking systems are better suited for action representation, *ii)* which features can be extracted from sensory data to reduce the dimensionality and increase the discriminativeness, and *iii)* how the similarity between actions can be measured.

## A. Whole-body Action Representation

Modern motion tracking systems adopt a kinematic model of the human body, the so-called skeletal model, consisting of a certain number of links connected by joints. The raw information available from the tracking system is a time series of skeletal poses sampled at different time instants. A possible way to represent whole-body actions is to collect a set of 3D joint positions sampled at different times, i.e. a set of Cartesian trajectories. As discussed in Section II, Cartesian trajectories depend on the reference frame in which the motion is expressed and on the length of human limbs. On the other hand, joint angles between two connected links in the skeletal model are naturally invariant to roto-translations and scaling factors. Hence, in this work, we represent an action as a set of joint angles trajectories, i.e. as the $J \times T$ matrix $\boldsymbol{A} = [\boldsymbol{a}_1, \ldots, \boldsymbol{a}_J]$, where $\boldsymbol{a}_j = \{a_j^t\}_{t=1}^T$ is the trajectory of the $j$-th joint angle, $J$ is the number of joints and $T$ is the number of time frames. One possibility is to directly use the raw time series $\boldsymbol{A}$ for action classification. Alternatively, as in this work, one can extract from $\boldsymbol{A}$ a feature vector (action descriptor) whose objective is to reduce the size of the input data and increase their discriminativeness.

## B. Coordination-Based Action Descriptor

The proposed action descriptor is based on two assumptions. The first assumption is that, while each subject can perform the same action in different manners generating different joint trajectories, all the subjects tend to activate the same set of joints [5]. For example, in a clapping action the arm joints are the most informative, while the rest are practically unused. The second assumption is that humans move the joints in a coordinated fashion [1], and, therefore, motion coordination is discriminative for motion recognition.

Building upon these assumptions, we define the CODE action descriptor $\mathcal{A}$ as the 5-tuple

$$\mathcal{A} \triangleq (I_m, \hat{\boldsymbol{\sigma}}, \hat{\boldsymbol{v}}_{\max}, \hat{\boldsymbol{v}}_{\min}, \boldsymbol{c}) \quad (1)$$

where $I_m$ contains the indexes of the $J_m$ most informative joints (MIJ), $\hat{\boldsymbol{\sigma}} \in \mathbb{R}^{J_m}$, $\hat{\boldsymbol{v}}_{\max} \in \mathbb{R}^{J_m}$ and $\hat{\boldsymbol{v}}_{\min} \in \mathbb{R}^{J_m}$ are respectively the normalized variances, maximum and minimum velocities of the MIJ. The vector $\boldsymbol{c}$ is the correlation between each pair of MIJ and has $J_m(J_m - 1)/2$ components. In more detail, the vector $\boldsymbol{c}$ is obtained by concatenating the correlation coefficients $c_{ij}$, where $(i, j)$ is a couple of most informative joints of an action $A$. If an action has $J_m$ most informative joints, we will have $J_m(J_m - 1)/2$ pairwise combination. With the symbols $\mathcal{A}$ we denote a finite ordered list of elements (a tuple). Each element of this tuple is a vector. For implementation purposes, the elements of the 5-tuple $\mathcal{A}$ are stacked into an array of $N_C = J_m(J_m + 7)/2$. components. Hence, the number of MIJ $J_m$ determines the size of the descriptor and it has to be chosen in order to guarantee a good compromise between dimensionality (computation time) and recognition performance. Details about the action descriptor in (1) are provided in the rest of this Section.

*1) Selecting the Most Informative Joints:* During the execution of an action, not all the joints contribute in the same manner. Hence, a possible way to represent a motion is to find which joints contribute the most to the whole motion, i.e. which are the most informative joints (MIJ). The variance $\sigma_j, j = 1, \ldots, J$ of each joint angle trajectory is used to identify the $J_m \leq J$ most informative joints, considering that the higher the variance, the higher the contribution of that joint to the whole-body motion [5].

For a given action $\boldsymbol{A} = [\boldsymbol{a}_1, \ldots, \boldsymbol{a}_J]$, the variance is computed for all the $J$ columns of $\boldsymbol{A}$, obtaining the vector $\boldsymbol{\sigma}^a = [\sigma_1^a, \ldots, \sigma_J^a]^T$. The elements of $\boldsymbol{\sigma}^a$ are sorted as

$$\begin{aligned} (\boldsymbol{\sigma}^s, I^s) &= \text{sort}(\boldsymbol{\sigma}^a), \\ \boldsymbol{\sigma}^s &= [\sigma_1^s, \sigma_2^s, \ldots, \sigma_{J_m}^s, \ldots, \sigma_J^s]^T, \\ I^s &= \{i_1^s, i_2^s, \ldots, i_{J_m}^s, \ldots, i_J^s\} \end{aligned} \quad (2)$$

where the function $\text{sort}(\boldsymbol{u})$ sorts the elements of $\boldsymbol{u}$ in descending order and returns the sorted indexes $I^s$. The vector of normalized variances $\hat{\boldsymbol{\sigma}}$ of the $J_m$ MIJ is computed as

$$\begin{aligned} I_m &= \{i_1^s, i_2^s, \ldots, i_{J_m}^s\}, \\ \boldsymbol{\sigma} &= [\sigma_1^s, \sigma_2^s, \ldots, \sigma_{J_m}^s]^T, \\ \hat{\boldsymbol{\sigma}} &= \frac{\boldsymbol{\sigma}}{\sum_{j=1}^{J_m} \sigma_j^s} = [\hat{\sigma}_1, \hat{\sigma}_2, \ldots, \hat{\sigma}_{J_m}]^T \end{aligned} \quad (3)$$

The last expression in (3) guarantees that $\sum_{j=1}^{J_m} \hat{\sigma}_j = 1$. It is worth noticing that taking the variance of the MIJ $\hat{\boldsymbol{\sigma}}$ as action descriptor significantly reduces the amount of data. Indeed, as discussed in Section III-A, raw sensory data are $T \times J$ matrices, where $T$ is usually bigger than $J$, while $\hat{\boldsymbol{\sigma}}$ is a vector with $J_m \leq J$ components. In this work, we set $J_m = 20$, as motivated in Section IV-B.

The colormaps in Fig. 2 represent the normalized joint angle variances $\hat{\boldsymbol{\sigma}}^a = \boldsymbol{\sigma}^a / \sum_j^J \sigma_j^a$ as a function of the joint angle index. Three action classes are considered from the HDM05 database: *clap1Reps*, *clapAboveHead1Reps* and *squat1Reps*. Each action is repeated 5 times, and each repetition is associated to a repetition number. Let us firstly focus on a single action class, e.g. *squat1Reps* in Fig. 2(a). Each row of the colormap represents a repetition of *squat1Reps*. We can see that only a small subset of joints have not negligible variance and all the repetition have a common set of informative joints. Moreover, in Fig. 2, the class *clap1Reps* is compared, in terms of joint angle variances, with *squat1Reps* in Fig. 2(a) and with *clapAboveHead1Reps* in Fig. 2(b). Looking at the figure, it is evident how actions that use different MIJ, such as *squat1Reps* and *clap1Reps*, present a different joint variance pattern (see Fig. 2(a)). On the other hand, classes like *clapAboveHead1Reps* and *clap1Reps*, which have similar MIJ, present a similar variance pattern, as shown in Fig. 2(b).

MIJ can easily discriminate actions executed with different joints. Nevertheless, when dealing with large datasets, different classes with similar MIJ can become very common. To increase the discriminativeness, we enrich our descriptor with velocities and pairwise correlations between the MIJ.

*2) Maximum and Minimum Velocity of the MIJ:* The variance captures information on joint angular motion without

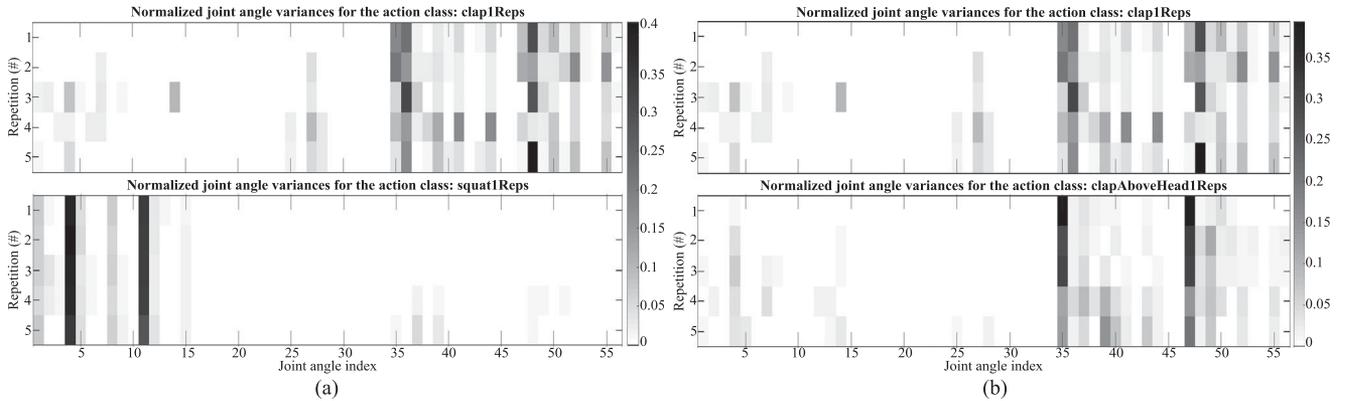

Fig. 2. Joint angle variances as a function of joint index and repetition number. (a) *clap1Reps* and *squat1Reps* have different sets of most informative joints. This kind of actions can be correctly classified considering only the relevant joints. (b) *clap1Reps* and *clapAboveHead1Reps* have similar sets of most informative joints. For this kind of actions misclassification may occur if only the most informative joints are considered as features.

considering the direction of the motion. Distinguishing between positive and negative joint rotations increases the informativeness of the descriptor and improves the recognition performance. The normalized maximum and minimum MIJ velocities

$$\hat{v}_{\max} = \frac{v_{\max}}{\sum_{j=1}^{J_m} |v_{max,j}|}, \quad \hat{v}_{\min} = \frac{v_{\min}}{\sum_{j=1}^{J_m} |v_{min,j}|} \quad (4)$$

are also considered in our descriptor. By construction, $\hat{v}_{\max}$ and $\hat{v}_{\min}$ are vectors with $J_m$ components.

*3) Pairwise Correlation of the MIJ:* Neuromechanical evidences show a certain degree of correlation between the most informative joints (or a subset of MIJ) [1], [2]. To exploit such a correlation, we enrich the descriptor with the vector $c$ of pairwise correlations of the $J_m$ most informative joints. In particular, given a MIJ trajectory $A_m = [a_1, \ldots, a_{J_m}] \in \mathbb{R}^{T \times J_m}$, one can compute the pairwise correlation matrix

$$C = \begin{bmatrix} 1 & c_{1,2} & \cdots & c_{1,J_m} \\ c_{2,1} & 1 & \cdots & c_{2,J_m} \\ \vdots & \vdots & \ddots & \vdots \\ c_{J_m,1} & c_{J_m,2} & \cdots & 1 \end{bmatrix} \quad (5)$$

where the element $-1 \leq c_{ij} \leq 1$ represents the linear correlation between the joint $i$ and $j$ and it is computed as

$$c_{ij} = \frac{\sum_{t=1}^{T}(a_i^t - \bar{a}_i)(a_j^t - \bar{a}_j)}{\sqrt{\sigma_i^s}\sqrt{\sigma_j^s}} = \frac{\text{cov}(a_i, a_j)}{\sqrt{\sigma_i^s}\sqrt{\sigma_j^s}} \quad (6)$$

The quantities $\bar{a}_i$ and $\bar{a}_j$ in (6) are the mean values of $a_i$ and $a_j$ respectively, while the variances $\sigma_i^s$ and $\sigma_j^s$ are defined as in (3). The numerator of (6) represents the covariance between $a_i$ and $a_j$. By construction, the correlation matrix $C$ in (5) is symmetric with unitary diagonal elements. The $J_m(J_m - 1)/2$ different entries in $C$ are stacked into the correlation vector $c$ and used to augment our descriptor. The procedure to compute CODE is summarized in Algorithm 1.

### C. Analysis of Space and Time Complexity

We report in Table I the (computational) time and space complexity of the CODE descriptor as a function of the number of most informative joints $J_m$ and the number of action time

**Algorithm 1:** CODE Descriptor.

**input:** Action matrix $A$, MIJ number $J_m$
1: Compute normalized variance and MIJ indexes
  $\sigma^a = $ variance $(A)$
  $(\sigma^s, I^s) = $ sort $(\sigma^a)$
  $\sigma = [\sigma_1^s, \sigma_2^s, \ldots, \sigma_{J_m}^s]^T$
  $I_m = \{i_1^s, i_2^s, \ldots, i_{J_m}^s\}$
  $\hat{\sigma} = \sigma / \sum_{j=1}^{J_m} \sigma_j^s$
2: Compute normalized velocities
  $\hat{v}_{\max} = v_{\max} / \sum_{j=1}^{J_m} v_{max,j}$
  $\hat{v}_{\min} = v_{\min} / \sum_{j=1}^{J_m} v_{min,j}$
3: Compute correlation vector
  $C = \{c_{ij}\}_{i=1,j=1}^{i=J_m,j=J_m}$, where
  $c_{ij} = \text{cov}(a_i, a_j)/(\sqrt{\sigma_i^s}\sqrt{\sigma_j^s})$
  stack the upper (or lower) triangular part of $C$ into the vector $c$
4: **return** $[I_m, \hat{\sigma}, \hat{v}_{\max}, \hat{v}_{\min}, c]$

TABLE I
TIME AND SPACE COMPLEXITY OF CODE AS A FUNCTION OF THE NUMBER OF MIJ $J_m$ AND THE NUMBER OF FRAMES $T$

|  | Time Complexity | Space Complexity |
| --- | --- | --- |
| MIJ number ($J_m$) | $\mathcal{O}(J_m^2)$ | $\mathcal{O}(J_m^2)$ |
| Frames ($T$) | $\mathcal{O}(T)$ | $\mathcal{O}(1)$ |
| Overall ($J_m, T$) | $\mathcal{O}(J_m^2 T)$ | $\mathcal{O}(J_m^2)$ |

frames $T$. As described previously in this section, CODE has $J_m(J_m + 7)/2$ components. Hence, using the big O notation [23], its space complexity is $\mathcal{O}(J_m^2)$. The space complexity is $\mathcal{O}(1)$, since the size of CODE is independent from the number of time frames $T$. Regarding the time complexity as a function of $J_m$, the most time-complex operation in Algorithm 1 is step 3, i.e., computation of the correlation vector. The computation of the correlation coefficient is performed as in (6) for each pair of MIJ. Since there are $J_m(J_m - 1)/2$ combinations of MIJ pairs, the time complexity as a function of $J_m$ is $\mathcal{O}(J_m^2)$. The time complexity as a function of the number of time frames is $\mathcal{O}(T)$, since the computation of variances in (3), the computation of normalized velocities in (4), and the computation of the correlation vector in (6) have $\mathcal{O}(T)$ time complexity. Overall, CODE has $\mathcal{O}(J_m^2 T)$ time complexity and $\mathcal{O}(J_m^2)$ space complexity.

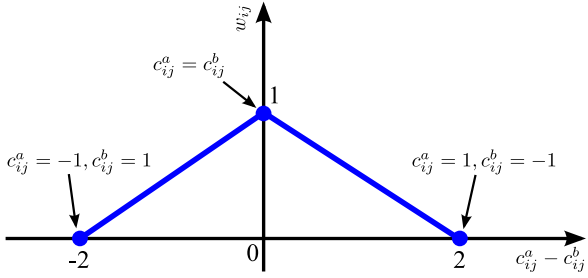

Fig. 3. Value of the weight $w_{ij}$ as a function of $c^a_{ij} - c^b_{ij}$.

TABLE II
DATASETS CHARACTERISTICS

| Dataset | Subjects (#) | Classes (#) | Actions (#) | Frame Rate (Hz) |
|---|---|---|---|---|
| HDM05 | 5 | 80 | 2337 | 120 |
| R-HDM05 | 5 | 16 | 401 | 120 |
| MHAD | 12 | 11 | 659 | 480 |

### D. Correlation-Based Similarity Measure

As described in Section III-B, CODE represents an action with a vector of dimension $N_C$. To measure the similarity among actions, we propose a novel similarity measure called Correlation-based Similarity Measure (CSM).

Consider the two action descriptors $\mathcal{A}^a$ and $\mathcal{A}^b$ where $\mathcal{A}^u = (I_m^u, \hat{\boldsymbol{\sigma}}^u, \hat{\boldsymbol{v}}^u_{\max}, \hat{\boldsymbol{v}}^u_{\min}, \boldsymbol{c}^u)$, $u = a, b$. Let us define the set $\mathcal{S} = \{(i,j) \in I_m^a \cap I_m^b | i \neq j\}$. In practice, $\mathcal{S}$ contains the pairs of MIJ that are common to $\mathcal{A}^a$ and $\mathcal{A}^b$. The CSM between two action descriptors $\mathcal{A}^a$ and $\mathcal{A}^b$ is defined as

$$CSM(\mathcal{A}^a, \mathcal{A}^b) = \sum_{i,j \in \mathcal{S}} w_{ij}[(\hat{\sigma}^a_i + \hat{\sigma}^a_j + \hat{\sigma}^b_i + \hat{\sigma}^b_j) \\ + (\hat{v}^a_{max,i} + \hat{v}^a_{max,j} + \hat{v}^b_{max,i} + \hat{v}^b_{max,j}) \\ + (\hat{v}^a_{min,i} + \hat{v}^a_{min,j} + \hat{v}^b_{min,i} + \hat{v}^b_{min,j})] \quad (7)$$

where the weight $w_{ij} = 1 - 0.5|c^a_{ij} - c^b_{ij}|$ is maximum ($w_{ij} = 1$) when the action $a$ and $b$ have the same correlation between the common most informative joints $i$ and $j$. The weight $w_{ij}$ is minimum ($w_{ij} = 0$) if the common MIJ $i$ and $j$ are perfectly correlated in action $a$ ($c^a_{ij} = 1$) and anti-correlated in action $b$ ($c^b_{ij} = -1$), or viceversa ($c^a_{ij} = -1$ and $c^b_{ij} = 1$). The correlation-based similarity measure in (7) is a summation of variances and velocities of common MIJ weighted by the differences in pairwise correlations between the two actions. Hence, two actions which use the same MIJ, but are characterized by a different correlation pattern, will have a low CSM score. High values of CMS indicate a high similarity between two actions. CSM is zero if two actions have no common MIJ or if all the MIJ are anti-correlated. Moreover, the joints that present a higher variance, maximum and minimum velocities give more contribution to the evaluation of similarity CSM than joint with low variance, and velocities. Fig. 3 shows the value of the weight $w_{ij}$ for two actions $a$ and $b$ as a function of the difference in correlation between two common most informative joints $i$ and $j$.

## IV. EXPERIMENTAL RESULTS

In order to prove the effectiveness of our approach, we perform three types of experiments on the public motion datasets HDM05 [6] and MHAD [7]. In the first type of experiments, we evaluate the accuracy on the whole HDM05 dataset as a function of the number of most informative joints with different features and different similarity measures. In the second set of experiments, we evaluate accuracy, precision and recall of CODE. The third class of experiments consists in a comparison with other descriptors in the literature. In order to reduce high-frequency noise, we apply a butterworth filter with cut-off frequency of 10 Hz.

### A. Dataset Description

We use three different datasets for our experiments: (i) HDM05, (ii) Reduced HDM05 and (iii) MHAD. The main characteristics of each dataset are summarized in Table II.

The HDM05 dataset contains 2337 actions split into 130 classes, and the actions are performed by 5 subjects. We consider 80 classes obtained by merging the motion recordings that contain multiple executions of the same action. For example, clap one repetition and clap five repetitions have been considered to be in the same class.

The Reduced HDM05 (R-HDM05) dataset is a subset of HDM05 composed by 401 action sequences split into the 16 classes: "emphdepositFloorR (1), elbowToKnee1RepsLelbowStart (2), grabHighR (3), hopBothLegs1hops (4), jogOnPlaceStartAir2StepsLStart (5), jumpDown (6), jumpingJack1Reps (7), kickLFront1Reps (8), lieDownFloor (9), rotateArmsBothBackward1Reps (10), sitDownChair (11), sneak2StepsLStart (12), squat1Reps (13), standUpKneelToStand (14), throwBasketball (15), throwFarR (16)". The numbers in brackets are the class labels used in Fig. 6. These are the action classes chosen in [5], which we adopt to perform comparisons.

MHAD is constituted by 11 classes: "jumping (1), jumping jacks (2), bending (3), punching (4), waving two hands (5), waving one hand (6), clapping (7), throwing (8), sit down (9), stand up (10), sit down/stand up (11)". The numbers in brackets are class labels used in Fig. 7. Each action is performed by 12 subjects 5 times, yielding a total of 659 actions (1 erroneous action was removed from the database).

### B. Number of Most Informative Joints

The goal of this experiment is two-fold. First, it shows the contribution of the different CODE components in Section III-B.

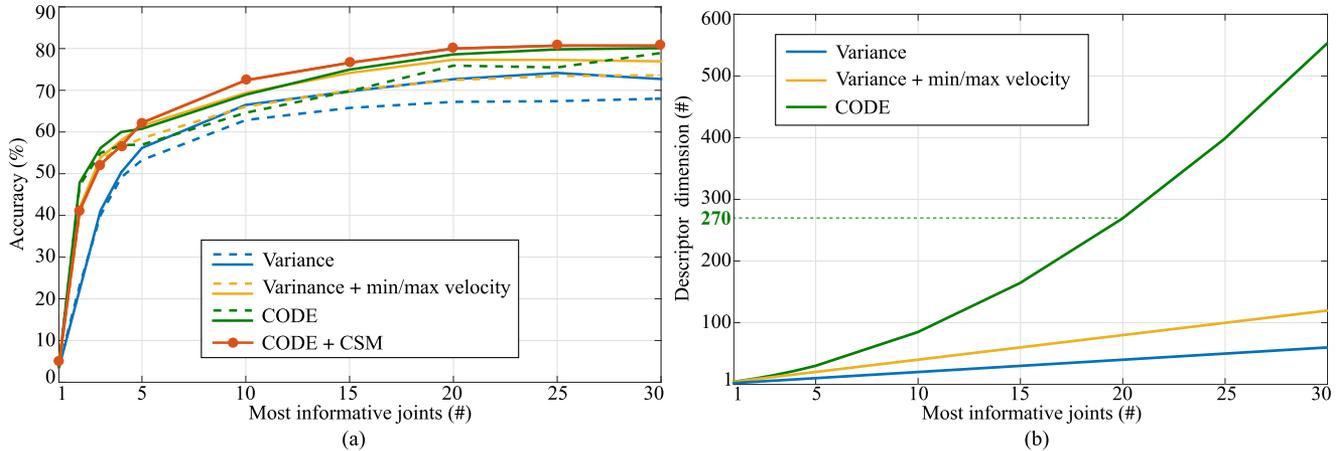

Fig. 4. Results on the HDM05 dataset (2337 actions and 80 classes). (a) Recognition results for different values of $J_m$ and different features vectors. (b) Motion descriptors that consider only variance or variance and velocity as features grow linearly with $J_m$, while CODE grows quadratically. CODE with CSM offers a good compromise between recognition rate (80.0%) and descriptor dimension (270 components with $J_m = 20$). (a) Accuracy as a function of MIJ number. Dashed lines are obtained with Euclidean distance, solid lines with Manhattan distance. (b) Descriptors dimension as a function of MIJ number.

Second, it investigates how to choose an efficient number of most informative joints. To guarantee a statistical relevance, we tried CODE on a large set of actions and classes, i.e., the 80 classes and 2337 actions of HDM05. The accuracy of CODE, evaluated as a function of the Most Informative Joints (MIJ) number $J_m$, is shown in Fig. 4(a). The accuracy is computed as the ratio between the number of total test inputs correctly classified and the number of test inputs. In the figure, CODE with the proposed CSM is compared with descriptors based (i) only on variance of MIJ, (ii) on variance and joint angular velocities of MIJ, (iii) on variance, velocity, and correlation of MIJ. The results show that all CODE features contribute to improve the recognition rate.

The continuous lines in Fig. 4(a) denote the use of Manhattan distance, while the dashed lines denote Euclidean distance to evaluate the similarity between actions. In case of CODE+CSM, we use our proposed metrics to evaluate the similarity. We can see that, in general, Manhattan distance performs better than Euclidean, and CSM performs better than Manhattan distance for $J_m \geq 5$. An advantage of the proposed Correlation-based Similarity Measure is that CODE+CSM performs better with less MIJ with respect to Euclidean and Manhattan distances. For example, with $J_m = 20$, CODE+CSM achieves 80.0% of accuracy, while CODE+Manhattan achieves 78.6% of accuracy. When increasing the number of MIJ ($J_m \geq 20$), the difference between the metrics becomes smaller. For example, with $J_m = 30$, CODE+CSM achieves 80.7% of accuracy, while CODE+Manhattan achieves 80.1% of accuracy. We can conclude that CSM achieves better performance than Euclidean and Manhattan distances with a reduced number of MIJ.

Fig. 4(b) shows the dimension of CODE as a function of the number of most informative joints. The dimension of CODE increases quadratically with $J_m$. This is an expected result considering the spatial complexity analysis in Section III-C. Using only variance and variance+velocities, the size of the descriptor increases linearly. The price paid for a more precise characterization of the motion is an increase in the descriptor dimensionality. Considering the accuracy in Fig. 4(a) (80.0% with $J_m = 20$ and 80.7% with $J_m = 30$) and the descriptor size in Fig. 4(b) (270 components with $J_m = 20$ and 555 components with $J_m = 30$), we can conclude that CODE+CSM with $J_m = 20$ offers a good compromise between recognition rate and size of the descriptor.

TABLE III
10-FOLD CROSS-VALIDATED RESULTS WITH CODE+CSM

| Dataset | Accuracy (%) (mean ± std) | Precision (%) (mean ± std) | Recall (%) (mean ± std) | Time (s) (mean ± std) |
|---|---|---|---|---|
| MHAD | 96.4 ± 2.9 | 96.7 ± 3.3 | 96.8 ± 2.5 | 9.54 ± 0.61 |
| R-HDM05 | 96.0 ± 2.7 | 94.5 ± 3.5 | 95.6 ± 3.8 | 0.64 ± 0.02 |
| HDM05 | 80.0 ± 2.9 | 73.7 ± 2.6 | 73.0 ± 2.7 | 3.84 ± 0.1 |

C. Performance Evaluation

Using 10-fold cross-validation, accuracy, precision, and recall of CODE have been evaluated on three datasets: HMD05, R-HMD05, and MHAD. Precision is obtained as the ratio between true positives and the sum of true positives and false positives. Recall is obtained as the ratio between true positives and the sum of true positives and false negatives. Also, we report the time to compute CODE for all the actions of each dataset. The computer used for the evaluation has an Intel Core $i7 - 4790$ K - 4 Cores CPU, and 16 GB of memory. CODE is implemented in Matlab 2014b. The results, summarized in Table III, are obtained using CODE with CSM, $J_m = 20$ and 1-NN classification. The average accuracy of CODE on HDM05 is 80.0%, precision is 73.7% and recall is 73.0%. The time to compute the CODE for all the actions of HDM05 is 3.84 s with our unoptimized Matlab implementation. For the R-HDM05 dataset, we achieve the average accuracy of 96.0%, the average precision of 94.5%, and the average recall of 95.6%. The time to compute the descriptor for all action of R-HDM05 is 0.64 s. In the experiments on the MHAD dataset accuracy, precision, and recall are 96.4%, 96.7% and 96.8%, respectively, while the time to compute CODE for all the actions is 9.54 s. In Fig. 5, the robustness of CODE in presence of Additive Gaussian White Noise (AGWN) is evaluated. We corrupted the joint angle signals with AGWN of standard deviation in the range [0, 5] deg. For R-HDM05, with a standard deviation of 5 deg the accuracy is 93.8%, for MHAD is 93.3%, while for HMD05 the accuracy is 77.5%. Roughly, we loose about 3% accuracy corrupting the signals with additional AGWN of 5 deg standard deviation.

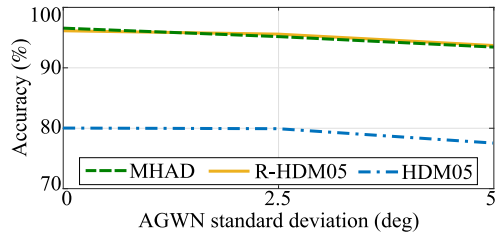

Fig. 5. Accuracy of CODE for different values of the AGWN standard deviation, evaluated on HDM05, R-HDM05, and MHAD.

TABLE IV
CLASSIFICATION RESULTS FOR THE R-HDM05 DATASET

| Descriptor | Classification | Accuracy (%) |
|---|---|---|
| CODE + CSM | 1-NN | **98.4** |
| SMIJ [5] | 1-NN | 91.5 |
| HMIJ [5] | 1-NN | 73.5 |
| HMW [5] | 1-NN | 77.4 |
| LDSP [5], [22] | 1-NN | 67.8 |

### D. Comparison With Angle-Based Approaches

We compare the recognition performance of CODE with the state-of-the-art descriptors in [5], [16], [22]. The comparison is carried out on both the R-HDM05 and the MHAD datasets. As in the previous experiments, we use CSM to measure the similarity between the CODE descriptors of different actions and $J_m = 20$ most informative joints.

*R-HDM05:* For a fair comparison, we adopt the same 16 classes (see Section IV-A) and the same cross-subject validation protocol used in [5]. In particular, we consider 3 subjects (219 action sequences) for training and the remaining 2 subjects (182 action sequences) for testing. Cross-subject validation is particularly interesting to demonstrate the generalization capabilities of CODE across different users. Additionally, we compare CODE with Histograms of Most Informative Joints (HMIJ) [5], Histogram-of-Motion Words (HMW) [5], and Linear Dynamical System Parameter (LDSP) [22]. The results of this comparison are shown in Table IV. We can see that the best results are achieved by CODE, with an accuracy of 98.4%. The confusion matrix relative to this case study in presented in Fig. 6. The actions that do not achieve 100.0% accuracy are *jumpDown* (6), *kickLFront1Reps* (8), *lieDownFloor* (9). The action *jumpDown* has 87.0% accuracy and is confused with *hopBothLegs1hops* (4) in 13.0% of cases. The accuracy for *kickLFront1Reps* (8) is 92.0% and it is confused with *jumpDown* (6). *lieDownFloor* (9), which presents an accuracy of 90.0%, is confused with *jumpDown* (6) in the 10.0% of cases.

*MHAD:* The comparison between CODE, SMIJ, HMIJ, and LDSP on the classes of the MHAD database is reported in

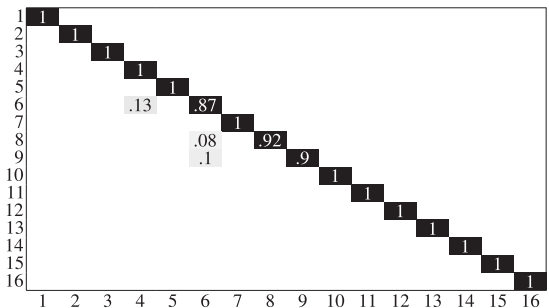

Fig. 6. Confusion matrix for 1-NN classification of the R-HDM05 dataset.

TABLE V
CLASSIFICATION RESULTS FOR THE MHAD DATASET

| Descriptor | Classification | Accuracy (%) |
|---|---|---|
| CODE + CSM | 1-NN | **98.5** |
| SMIJ [5] | 1-NN | 94.5 |
| HMIJ [5] | 1-NN | 80.3 |
| HMW [5] | 1-NN | 77.7 |
| LDSP [5], [22] | 1-NN | 84.9 |

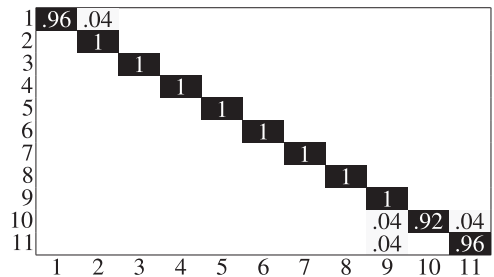

Fig. 7. Confusion matrix for 1-NN classification of the MHAD dataset.

Table V. In this case, CODE achieves 98.5% accuracy and the second best is SMIJ that achieves 94.5%. In this experiment, 7 subjects are chosen for training (384 action sequences) and 5 (275 action sequences) for testing, according to the cross-subject validation protocol adopted in [5]. The confusion matrix is shown in Fig. 7. We can see that the accuracy of CODE is 100.0% for the majority of the classes, except for three classes: *jumping* (1), *sit down* (10), *sit down and stand up* (11). The accuracy is 96.0% for the action *jumping* (1), which has been confused in 4.0% of cases with the action *jumping jacks* (2). Moreover, the action *sit down* (10) presents a recognition rate of 92.0%, since it is confused in 4.0% of cases with *stand up* (9), and in 4.0% of cases with *sit down and stand up* (11). The action *sit down and stand up* achieves 96.0% accuracy and it is confused with *sit down* in 4.0% of cases.

### E. Comparison With Position-Based Approaches

In addition to the comparison with angle-based methods, we compare CODE also with approaches that use joint Cartesian positions. The two representations work with different input data, i.e. joint angles and 3D joint positions, respectively. Since the recognition performance strongly depends on the type of input data, the comparisons in terms of accuracy are merely indicative.

However, the scope of this section is to discuss basic differences between CODE and most successful position-based recognition approaches. First, we compared CODE with the skeleton quad descriptor presented in [16]. This approach obtains 93.89% on a subset (11 classes) of the R-HDM05 dataset. On the same subset, CODE achieves 100% accuracy. The second comparison is with the template-based approach (TBA) presented in [14]. It adopts DTW [15] to align the training trajectories with the test trajectories and has been tested with 9 classes [14] of HDM05 dataset, achieving 98.0% accuracy. On the same classes CODE achieves 98.3% accuracy. In terms of accuracy, the performance of TBA and CODE are similar on the tested classes. However, TBA has a $\mathcal{O}(T)$ spatial complexity (to store the entire joint position trajectories) and $\mathcal{O}(T^2)$ time complexity (to align training and test trajectories with DTW), while CODE has $\mathcal{O}(1)$ spatial complexity and $\mathcal{O}(T)$ time complexity (see Table I). The third comparison is with the skeleton-based approach (SKA) in [10]. It uses a deep neural network and a frame-by-frame classification to recognize motion capture sequences. The experiments are performed on 2337 actions of HDM05 split in 65 classes, achieving 95.6% accuracy. On the same action set CODE achieves 87.7% accuracy. In terms of accuracy SKA performs better than CODE. However, SKA uses a more complex descriptor with $33 \times T$ elements, where $T$ is the number of time frames. The space complexity is therefore $\mathcal{O}(T)$, while CODE has a fixed size of $270 \times 1$ elements. Moreover, SKA adopts a classification algorithm based on deep learning, which requires a relatively long training time, while in this work we use a 1-NN classifier to keep the system simple and fast, according to the requirements typical of robotic systems.

## V. Conclusion and Future Work

In this work, we presented CODE, a COordination-based action DEscriptor. CODE is based on the assumption, accepted in neuromechanics, that humans move in a coordinated fashion. CODE encodes the coordination properties of human motion by computing the pairwise correlations between the most informative joints. With experiments on two different datasets containing a large set of actions, we have shown that, including information about correlation and about joint velocities, the recognition performance improves significantly. The size of CODE is independent from the action duration and increases quadratically with the number of most informative joints. The comparisons showed that CODE outperforms several approaches for action recognition.

Future work will consist in evaluating CODE on representations based on Cartesian joint positions. Most renowned works in neuromechanics, in fact, discuss human motion correlation at a joint angle level. Therefore, the possibility to encode joint Cartesian positions with CODE-like descriptors requires further investigation. In this work, CODE is used only for action classification. In order to segment streams of data before the classification, CODE can be combined with a state-of-the-art segmentation method such as [24]. A future work direction will consist in applying the basic concept of CODE also to the segmentation problem.